\documentclass[sigconf]{acmart}

\AtBeginDocument{%
  \providecommand\BibTeX{{%
    \normalfont B\kern-0.5em{\scshape i\kern-0.25em b}\kern-0.8em\TeX}}}

\usepackage{graphicx}
\usepackage{subfigure}
\usepackage{amsfonts,latexsym}
\usepackage{amsmath}
\usepackage{microtype}
\usepackage{booktabs}
\usepackage{tabularx}
\usepackage{indentfirst} 
\usepackage{multicol}
\usepackage{pdfpages}
\usepackage{multirow}
\usepackage{listings}
\usepackage{xspace}

\newcommand\sysname{MedLeak\xspace}

\copyrightyear{2025}
\acmYear{2025}
\setcopyright{cc}
\setcctype{by}
\acmConference[CHASE '25]{ACM/IEEE International Conference on Connected
Health: Applications, Systems and Engineering Technologies}{June 24--26,
2025}{New York, NY, USA}
\acmBooktitle{ACM/IEEE International Conference on Connected Health:
Applications, Systems and Engineering Technologies (CHASE '25), June 24--26,
2025, New York, NY, USA}
\acmDOI{10.1145/3721201.3721375}
\acmISBN{979-8-4007-1539-6/2025/06}

\begin{document}
\title{\sysname: Multimodal Medical Data Leakage in Secure Federated Learning with Crafted Models}

\author{Shanghao Shi}
\affiliation{
  \institution{Virginia Tech}
  \city{Arlington, VA}
  \country{USA}}
  
\author{Md Shahedul Haque}
\affiliation{
  \institution{Virginia Tech}
  \city{Arlington, VA}
  \country{USA}}

\author{Abhijeet Parida}
\affiliation{
  \institution{Children's National Hospital}
  \city{Washington, D.C.}
  \country{USA}}
  
\author{Chaoyu Zhang}
\affiliation{
  \institution{Virginia Tech}
  \city{Arlington, VA}
  \country{USA}}

\author{Marius George Linguraru}
\affiliation{
  \institution{Children's National Hospital, \\ George Washington University}
  \city{Washington, D.C.}
  \country{USA}}

\author{Y. Thomas Hou}
\affiliation{
  \institution{Virginia Tech}
  \city{Blacksburg, VA}
  \country{USA}}

\author{Syed Muhammad Anwar}
\affiliation{
\institution{Children's National Hospital, \\ George Washington University}
\city{Washington, D.C.}
\country{USA}}

\author{Wenjing Lou}
\affiliation{
  \institution{Virginia Tech}
  \city{Arlington, VA}
  \country{USA}}

\renewcommand{\shortauthors}{Shanghao Shi, et al.}

\begin{abstract}

Federated learning (FL) allows participants to collaboratively train machine learning models while keeping their data private, making it ideal for collaborations among healthcare institutions on sensitive datasets.
However, in this paper, we demonstrate a novel privacy attack called \sysname, which allows a malicious participant who initiates the FL task as the server to recover high-quality site-specific private medical images and text records from the model updates uploaded by clients.
In \sysname, a malicious server introduces an adversarially crafted model during the FL training process. Honest clients, unaware of the insidious changes in the published model, continue to send back their updates as per the standard FL training protocol. Leveraging a novel analytical method, \sysname can efficiently recover private client data from the aggregated parameter updates. This recovery scheme is significantly more efficient than the state-of-the-art solutions, as it avoids the costly optimization process. Additionally, the scheme relies solely on the aggregated updates, thus rendering secure aggregation protocols ineffective, as they depend on the randomization of intermediate results for security while leaving the final aggregated results unaltered.

We implement \sysname on medical image datasets MedMNIST, COVIDx CXR-4, and Kaggle Brain Tumor MRI datasets, as well as the medical text dataset MedAbstract. The results demonstrate that the proposed privacy attack is highly effective on both image and text datasets, achieving high recovery rates and strong quantitative scores. 
We also thoroughly evaluate \sysname across different attack parameters, providing insights into key factors that influence attack performance and potential defenses. Furthermore, we perform downstream tasks, such as disease classification, using the recovered data, showing no significant performance degradation compared to the original training samples. Our findings validate the need for enhanced privacy measures in federated learning systems, particularly for safeguarding sensitive medical data against powerful model inversion attacks.

\end{abstract}




\maketitle 

\section{Introduction}

Federated learning (FL) has developed as a key enabling technology for the future implementation of AI-powered medical diagnosis and treatment systems \cite{brisimi2018federated, rieke2020future,sheller2020federated,adnan2022federated,pati2021federated,dayan2021federated,nguyen2022federated,ng2021federated}. FL allows medical centers to collaboratively train machine-learning models for various clinical tasks such as disease classification and clinical diagnosis, without sharing private patient information. This is crucial because medical centers are bound to preserve patient privacy and their data usage is strictly restricted in many clinical applications in accordance with regulatory guidelines.
Under the FL framework, distributed training is set up in a way where hospitals that own private clinical data usually serve as clients, and a server -- either hosted at one of the collaborating sites or maintained by a third party, integrates the model updates received from each client to orchestrate the federated learning paradigm. There are multiple open source (such as NVFLARE \cite{roth2022nvidia} and OpenFL \cite{reina2021openfl}) as well as commercial platforms (such as Rhino FCP \cite{stephens2021rhino}), designed to streamline the implementation of FL. During the FL training process, only the model updates, which refer to either the gradients or parameter updates, are exchanged between the participant and the server, while the private training samples are kept securely at the clinical site. Therefore, when first introduced, federated learning was considered to be privacy-preserving and the model updates are regarded as safe vectors that hide training samples' private information \cite{mcmahan2017communication,zhang2021survey,wei2021user,pfitzner2021federated,wang2022squeezing}.


\paragraph{\textbf{Existing Privacy Attacks}} Recent privacy attacks challenged the privacy-preserving property associated with FL. It is demonstrated that a curious or malicious parameter server can extract information related to the training samples such as their labels, membership information, and even the whole training sample using the model updates \cite{nasr2019comprehensive,fu2022label,luo2021feature,wang2019eavesdrop,zhu2019deep,zhao2020idlg,geiping2020inverting,yin2021see,fowl2021robbing}. Of particular interest, the model inversion attacks (MIAs) \cite{zhu2019deep,geiping2020inverting,yin2021see,fowl2021robbing,pasquini2022eluding, shi2023scale} are a type of privacy attack aiming to recover the original training images. 
They take the individual model updates provided by the clients as inputs and reverse them back to the local training samples. This could be detrimental in medical applications, where such an attack can reconstruct patient-specific data. Existing MIAs usually formulate this reverse process as an optimization problem and have been shown to achieve good recovery performance in recovering high-fidelity training images from the model updates, when enough optimization iterations are performed. This completely exposes the information that the FL system has been designed to protect. 

However, existing optimization-based MIAs are facing serious scalability and efficiency challenges. In practice, they need the server to consume extensive resources (usually hundreds of computation seconds with large memory) to recover only a few images, making them virtually impractical for real-world systems. Further, such MIAs can also be prevented by a specialized multi-party computation (MPC) mechanism named secure aggregation (SA) protocol \cite{bonawitz2017practical,pillutla2022robust,burkhalter2021rofl}, whose fundamental idea is to use various cryptography primitives (e.g., secret sharing) to mask individual model updates with random values but keep their summation identical to the pre-masked value. In this way, the FL system can proceed to the training process without exposing \textit{individual model updates}, preventing the MIAs from reversing them back to local training samples.



\paragraph{\textbf{Our Attack}} In this paper, we propose a novel attack name \sysname that addresses the limitation of the existing works. Our attack design makes it very practical, which could be detrimental to privacy in existing FL systems. \sysname is an efficient attack, capable of recovering hundreds of training samples in a batch from the victim client within just a few seconds. \sysname can also break the SA protocols as it can recover the training samples directly from the \textit{summation value} of the model updates even though the individual ones are cryptographically masked. In addition, \sysname can be accomplished within one FL training round, making it very practical and stealthy. To further evade detection, the attacker can launch the attack in the initial FL training rounds to avoid hurting the FL training performance. A greedy attacker can even launch the attack multiple times during continuous training rounds to harvest as many local training samples as possible. 

Technically, \sysname is a two-phase attack including the attack preparation phase and the sample recovery phase. In the first phase, the attacker adds an additional two-linear layer module in front of the original model architecture and initializes the module with customized parameters before sending it to the clients.
For the target victim, the attacker initializes the parameters of the two-layer module to form a ``linear leakage'' module with the help of an auxiliary dataset that has the same data format and distribution as the training samples. This ``linear leakage'' module is a powerful mathematical tool that can perfectly reverse its gradients back to its inputs, which are identical to the training samples because we place this module as the first component of the model architecture. For other clients, their two-layer modules' parameters are crafted to form a ``zero gradient'' module, aiming to zero out their gradients and model updates. By doing so the aggregated model update is identical to the model update of the victim because all the others are set to zero, rendering the SA protocols useless. 

As medical data is usually composed of both image and text records, we explore and extend \sysname's capability to recover both data modalities.
Particularly, the recovery of medical text records is largely ignored in the existing literature and we are proposing \textit{the first} work in this direction to our knowledge. 
Compared to medical images, the recovery of medical text records is more challenging because they are discrete natural language words with different paragraph lengths in the input space, resulting in completely different FL model architectures and parameter calculation methodologies. To address this challenge, we slightly modify and customize our attack to insert the malicious module after the embedding layer rather than at the front of the whole model to first launch the recovery word embedding vectors, and then further reverse these word embedding vectors back to input tokens (words).

\begin{table*}[t]
\caption{A comparison between the existing MIAs and \sysname with respect to their attack assumptions, efficiency, scales, capabilities, and attack generalizability.}
\centering
\begin{tabular}{ccccccc} 
\hline
\textbf{Attacks} & \textbf{Attack Model}\qquad &  \textbf{Attack Efficiency}\qquad  & \textbf{Scale}\qquad & \textbf{Break SA?}\qquad & \textbf{Recover Text?}\qquad & \textbf{Targeted?} \\
\hline
\cite{zhu2019deep,zhao2020idlg,geiping2020inverting,yin2021see,lu2022april,wen2022fishing} & Honest-but-curious & Low (Optimization-based) & $10^1$ & No & No (Image-only) & No \\
 \cite{pasquini2022eluding} & Malicious server & Low (Optimization-based) & $10^1$ & Yes & No (Image-only) & Yes \\
 \cite{fowl2021robbing,zhao2023loki} & Extra attack module & High (Mostly closed-form) & $10^2$ to $10^3$ & Yes & No (Image-only) & No \\
\midrule
\sysname & Extra attack module & High (Closed-form) & $10^2$ to $10^3$ & Yes & Yes & Yes\\
\hline
\end{tabular}
\vspace{-5pt}
\label{tab:existing_attacks}
\end{table*}

We evaluated \sysname on MedMNIST \cite{medmnistv2}, COVIDx CXR-4 \cite{Wang2020}, Kaggle Brain Tumor MRI \cite{brian-tumor-mri} datasets for medical images, and on the MedAbstract dataset \cite{MedAbstract} for medical text records.
For medical images, we evaluated our attack performance using the recovery rate, structural similarity (SSIM) score, peak signal-to-noise ratio (PSNR) score, and attack time. Our results show that \sysname achieves excellent performance on these datasets to recover hundreds of images simultaneously with high recovery rates and quantitative scores, with only a few seconds of execution time. On visual inspection, the recovered images are virtually indistinguishable from the original images. 
We compared the performance of \sysname with three existing MIAs. Our results demonstrate that \sysname achieves better quantitative scores and is comparatively much more efficient. We further fed the recovered images to downstream disease diagnosis tasks. Our results show that the recovered images achieved a classification performance close to the original images, validating the effectiveness of our attack. 
For medical text records, we evaluated our attack performance using the recovery rate, word error rate (WER), and attack time. The results show that \sysname can accurately recover tens of long clinical data paragraphs (e.g., descriptions of patients' health conditions) spanning up to hundreds of words simultaneously. This highlights \sysname's strong capability to recover non-image modality records effectively and efficiently.

In summary, in this paper we present the following contributions:

\begin{enumerate}
    \item We propose \sysname, a novel and powerful MIA that is capable of recovering high-quality local training samples in large batches using model updates from FL clients efficiently, even when state-of-the-art cryptography-based defense mechanisms such as secure aggregation are employed.     
    \item Our attack represents a fundamental and practical privacy vulnerability of the medical FL system as it compromises individual clients' privacy. \sysname can target both medical image and text data, demonstrating its broad applicability in the medical domain.
    \item We provide rigorous mathematical analysis and proof for our attack. Our proposed attack design is closed-form, hence avoids incurring any computation-intensive optimization and significantly reduces the computational costs when compared to existing MIAs.
    \item We implement \sysname on medical images and text datasets under different practical assumptions in the FL systems. The results show that our attack can nearly perfectly recover different types of local training samples of the target victim. 
\end{enumerate}



\section{Background}

\subsection{Federated Learning}
We consider for each training round $t$, there are $n$ clients denoted by $\mathcal{C}=\{c_1,c_2,\cdots,c_n\}$ to be selected by the parameter server $S$ to collaboratively train a global model $G= f_{\theta}: \mathcal{X \to Y}$, with each client $c_i$ holds a local dataset $D_i$. In detail, the parameter server $S$ first publishes the global model parameters $\theta^{t}$ to the clients. Then each client trains the received global model $G_t$ for $L^t_i$ local rounds over $D_i$ to generate its model update $\delta_i^t$. Note that when $L^t_i=1$, the model update $\delta_i^t$ can be replaced by the gradient $g_i^t$. After that, the client $c_i$ sends the model update $\delta_i^t$ back to the server $S$ and the server employs model aggregation (such as using the federated average (FedAVG) algorithm \cite{mcmahan2017communication}) to conduct the training process:
\begin{equation}
    \begin{aligned}
        \theta^{t+1}=\sum_{i=1}^{n}\alpha_i \delta^t_i,
    \end{aligned}
\end{equation}
where $\alpha_i$ is the weight assigned to client $c_i$. The summation of all weights $\{\alpha_i\}_{i:c_i\in\mathcal{C}}$ is 1 and can be adjusted according to the size of local datasets $D_i^t$ to avoid training bias. The server may also employ alternative aggregation strategies (such as FedSGD algorithm \cite{mcmahan2017communication}) for model aggregation. In the following text, we will omit the notation $t$, because our attack and analysis are all conducted in a single FL training round.

Federated learning has been widely used in the healthcare domain, enabling different healthcare providers to collaborate with each other towards accomplishing the training of machine learning models for both natural language processing and imaging tasks \cite{pfitzner2021federated,lin2021fednlp,liu2021federated}. Such clinical FL systems usually leverage the patients' private information as the local training datasets. These can be the patients' radiology scans, textual reports, and tabular records describing the patients' visits and health conditions. 
This information is considered to be highly private and sensitive and is protected by strict governance laws such as HIPPA and GDPR. 
The medical FL system is deployed to protect such privacy by ensuring that the sensitive data never leaves the firewalls within the healthcare sites during the model training process. However, herein we will demonstrate with \sysname that the privacy-preserving property of current FL systems is under challenge.

\subsection{Model Inversion Attacks}
The model inversion attacks (MIAs) take the individual model updates $\delta_i$ as the inputs and aim to reconstruct them back to the local datasets $D_i$ held by the clients. This reversion problem has been formalized as an optimization problem, represented as $\arg\min_{\hat{D_i}}[d(\nabla \hat{D_i}-\nabla D_i)]$, where $\hat{D_i}$ refer to randomly initialized dummy samples and $d()$ refers to a distance function such as the second norm distance. 
To solve this optimization problem, \cite{zhu2019deep} utilizes the L-BFGS optimizer to reconstruct the dummy samples iteratively step by step until reaching a good optimization point. It is further improved by an analytical method that aims to recover the ground-truth labels of dummy samples from the gradients \cite{zhao2020idlg}, which significantly eases the optimization task and helps accomplish better attack performance. Later works have improved the optimization tools and focus on recovering larger batches of images on more practical machine learning models such as the ResNet \cite{geiping2020inverting,yin2021see,lu2022april,wen2022fishing}. However, their recovery sizes are still restricted to the scale of tens, representing a scalability challenge. Moreover, all the aforementioned MIAs require costly iterative optimization methods in their design, which introduce a very large overhead to recover each batch of input images. Further, these existing methods also cannot bypass the current SA protocols.

\subsection{Secure Aggregation}
To enhance the privacy of the federated learning systems, Bonawitz et al. \cite{bonawitz2017practical} proposed a new type of specialized MPC mechanism named the secure aggregation (SA) protocols to fulfill an abstract function of masking individual model updates $\delta_i$ to $u_i$ with random bits, while keeping the summation of the masked values $\sum_{i=1}^{n}u_i$ identical to those of the pre-masked values $\sum_{i=1}^{n}\delta_i$. 
Therefore, despite variations of detailed cryptographic design, all SA protocols ensure that the server \textit{cannot} obtain the individual model updates $\delta_i$ to launch any model inversion attacks, but can proceed with the FL training process with the aggregated model update $\sum_{i=1}^{n}u_i$, which is identical to $\sum_{i=1}^{n}\delta_i$.
Since its initial introduction, the SA protocols have been continuously refined to incorporate other properties including communication efficiency, drop-out resilience \cite{bell2020secure, choi2020communication, guo2020v, kadhe2020fastsecagg}, and security against malicious clients \cite{pillutla2022robust, burkhalter2021rofl}, making it the current state-of-the-art privacy protection mechanism for FL systems.

\textbf{Secure Aggregation under Challenge:} Under the honest-but-curious attack model, the SA protocols have proven to be secure against various MIAs. However, recent works adopt a stronger attack model to assume a \textit{proactive} attacker that modifies the global model's parameters and even its architecture before publishing it to the clients. Under this assumption, \cite{pasquini2022eluding} proposed a novel attack that retrieves a target individual model update from the aggregated result, breaking the SA protocols. The fundamental idea is to craft the model parameters to adversarial models and distribute different adversarial models to different clients strategically. The adversarial models are crafted to ensure that only the model update of the victim client is preserved while all the others are zeroed out. The limitation of this attack is that it incorporates a costly optimization process in its design, which introduces too much attack overheads.
\cite{fowl2021robbing} proposed another attack method to add crafted modules before the original model architecture. These additional modules are crafted with delicate mathematical designs to ensure that the model gradients can be perfectly reversed back to inputs whenever the server receives any model updates.
The limitation of this attack is that the attacker cannot link the recovered images to their owners, which means the SA protocols still preserve a certain level of privacy, known as ``privacy by shuffling''.
Later, \cite{zhao2023loki} addressed this issue by designing a more complex adversarial module composed of convolutional and linear layers before the original model to identify the client associated with the recovered images. However, using a computer vision-related architecture prohibits the attack from being adopted to recover text records.

\subsection{Attack Summary and Comparison}
In Tab. \ref{tab:existing_attacks}, we summarize and compare the existing MIAs according to their attack assumptions, efficiency, scales, capabilities, and generability.
In our design, we adopt the active attacker assumption to break the SA protocol and simultaneously address various limitations of the existing works.
In particular, we aim to address the following three problems: 1) \textit{attack efficiency}-our attack shall not employ any computation-intensive optimization process to incur large overheads; 2) \textit{attack effectiveness}- our attack's capability of recovering high-quality local training samples from the \textit{aggregated model updates} and attribute them back to individual clients even when SA protocols are incorporated, and 3) \textit{attack generality}- our attack can be applied to traditional image recovery tasks as well as text recovery task (currently less explored). 


\section{Threat Model}

\begin{figure}[t]
    \centering
    \includegraphics[width=0.45\textwidth]{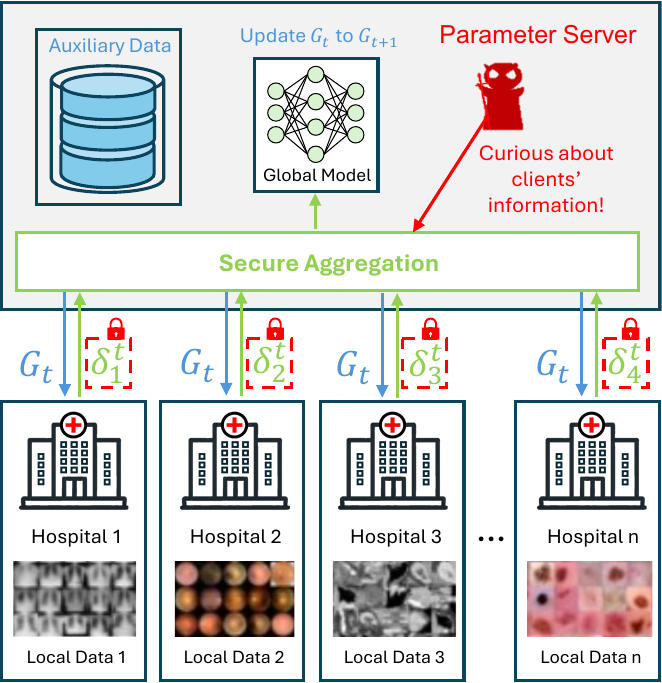}
    \caption{Threat Model. The server is considered to be a malicious attacker. The secure aggregation protocol is considered to be in place to protect the individual model updates.}
    \vspace{-10pt}
    \label{fig:threatmodel}  
\end{figure}

\begin{figure*}[t]
    \centering
    \includegraphics[width=0.96\textwidth]{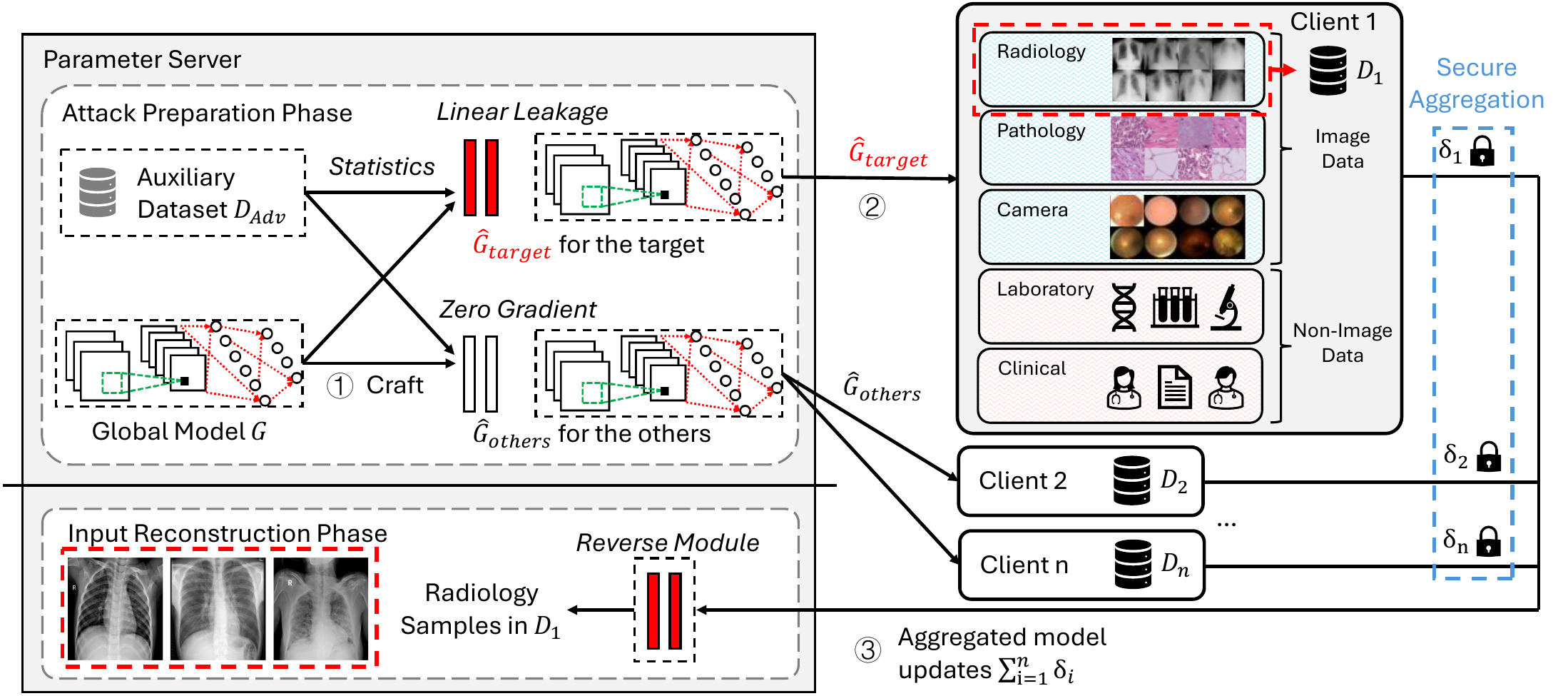}
    \caption{\sysname attack flow. 
    \sysname is a two-phase attack. In the first preparation phase, the attacker generates the adversarial global model. In the second reconstruction phase, the attacker sends the adversarial models to the clients and recovers the local samples when it receives their feedback. \sysname can reconstruct both image and non-image data and this figure demonstrates the reconstruction of the medical radiology images.
    }
    \vspace{-10pt}
    \label{fig:attackflow}  
\end{figure*}

In Fig.\ref{fig:threatmodel} we demonstrate the threat model of our attack. We consider the parameter server $S$ to be a malicious party that is curious about the training samples held by clients (e.g., private patient images or text records). We consider the parameter server to be a proactive attacker and can actively modify model parameters and architectures from $G$ to $\hat{G}$ to achieve the attack goals, following the same assumption as \cite{pasquini2022eluding,fowl2021robbing,zhao2023loki, shi2023scale}. We assume the communication channels between the server and clients are secured and all messages can be authenticated. We assume the state-of-the-art SA protocols are in place and the server can only get access to the \textit{aggregated model updates} $\sum_{i=1}^{n}\delta_i$ without knowing anything about the individual values $\delta_i$. We consider the attacker is able to collect or obtain an auxiliary dataset $D_{aux}$ that has the same data format and can represent the target dataset well. The attacker can leverage various online resources such as publicly available datasets, image searching tools, and image generative tools to fulfill this requirement. For our use case, the availability of public chest X-rays and medical text datasets makes this task trivial. The goal of the attacker is to recover the local data samples of a target client $c_{target}$ just from the aggregated model updates $\sum_{i=1}^{n}\delta_i$. This can be mathematically expressed as: $D_{target}=Reverse(\sum_{i=1}^{n}\delta_i, \hat{G})$.
\section{Attack Method}\label{Method}


\subsection{Attack Overview}

As we assume the attacker only possesses the already-masked aggregated model updates $\sum_{i=1}^{n}\delta_i$ under the protection of the SA protocol, it is very challenging to identify an end-to-end method that directly reverses the aggregated results back to local training samples $D_{target}$.
To address this, we decompose the complex recovery problem into two different distinct tasks including the \textit{individual model update retrieval} and \textit{efficient model update reversion} tasks. The first task aims to retrieve the individual model update of the victim from the aggregated model updates, thus breaking the SA protocol. The second task aims to reverse the individual model update back to local samples, achieving the recovery sub-problem.

To accomplish them, we require the attacker to place an additional two-linear-layer module $L_{adv}$ with the rectified linear unit (ReLU) activation function in between at the beginning of the original global model, i.e. $\hat{G}=G\oplus L_{adv}$.
The dimension of this module is identical to the image dimension. 
We require the attacker to initialize the parameters of the linear module $L_{adv}$ to form different adversarial modules and distribute them to different clients accordingly. For all the other clients except the victim, the attacker crafts the ``zero gradient'' modules $L_{zero}$ to ensure that the gradients and model updates of these clients are always zero, i.e. $\hat{G}_{others}=G\oplus L_{zero}$. By doing this, the attacker guarantees that only the model update of the victim client is exposed, accomplishing task one.
For the victim client, the attacker crafts a ``linear leakage'' module $L_{linear}$, aiming to reverse its model update back to local training samples efficiently, i.e. $\hat{G}_{target}=G\oplus L_{linear}$. This module requires an auxiliary dataset to help generate essential parameters and can ensure that samples are \textit{perfectly recovered} with a mathematical proof, accomplishing task two. 
Detailed attack flow and module designs are introduced in the next section.

\subsection{Detailed Attack Flow}

We demonstrate the attack flow in Fig.\ref{fig:attackflow}. \sysname is a two-phase attack including 1) attack preparation and 2) sample reconstruction phases. In the attack preparation phase, the attacker crafts the adversarial global model $\hat{G}$ and initializes it with different model parameters including $L_{zero}$ and $L_{linear}$ (step \textcircled{1}) before publishing them to different clients (step \textcircled{2}). Then in the second phase, the attacker collects the aggregated model updates and uses an analytical method to reverse it back to local training samples (step \textcircled{3}). \sysname can be used to reconstruct both medical image and non-image data. In the following parts, we will first introduce the overall attack flow and detailed design components via the image data reconstruction task, and then introduce how \sysname can be customized to accomplish the text reconstruction task.

\paragraph{\textbf{Attack Preparation:}} In this phase, the attacker crafts both the ``linear leakage'' module and ``zero gradient'' module once at the outset. Both modules require the estimation of essential parameters of a representative auxiliary dataset $D_{aux}$.
In detail, the attacker first estimates the cumulative density function (CDF) of the brightness feature $h(x)$ of the auxiliary dataset $D_{aux}$, denoted by $\psi(h(x))$, to represent the CDF of the local training dataset (which is unavailable), where $x$ refers to the input vector. After that, the attacker divides the distribution $\psi$ into equally $k$ bins by calculating $h_j=\psi^{-1}(j/k)$ where $j \in \{1,2,\cdots,k\}$, $\psi^{-1}$ refers to the inverse function of $\psi$, and $k$ equals to the neuron number of the first linear layer. 
By doing so, the brightness of a random input vector $x$ denoted by $h(x)$ will have the same probability of falling into each bin. This bin vector $\mathbf H=\left[h_1,h_2,\cdots,h_k \right]$ is the key vector to form both attack modules.

\textbf{Linear Leakage Module:}
Assuming the weight and bias matrix of the two-layer module $L_{adv}$ are $w_1, b_1$ and $w_2, b_2$ respectively.
For the target victim, the attacker initializes the ``linear leakage'' module $L_{linear}$ with the following steps: (1) having $w_1$'s row vectors all identical to $\left[\frac{1}{d},\frac{1}{d}, \cdots, \frac{1}{d}\right]$, where $d$ refers to the dimension of the input images, resulting in calculating the brightness feature on each neuron when the local training images are sent into the model during the FL training process; (2) having the bias $b_1$ identical the opposite value of $\mathbf{H}$, i.e.,$b_1=-\mathbf{H}$; (3) having all row vectors of $w_2$ the same. 

\textbf{Zero Gradient Module:}
The ``zero gradient'' module is initialized in the same way as the ``linear leakage'' module except in step (2), in which the attacker has the bias vector $b_1$ equal to $\mathbf{H'}=-\left[h_k, h_k, \cdots, h_k\right]$. By doing so, the output of the first linear layer will always be smaller than zero because $h_k$ is the largest possible brightness and all possible input $x$'s brightness is smaller than it. Considering we use the ReLU activation function after the first layer, the input to the second layer and the gradients of the first layer shall always be zero because of the ReLU function's mathematical property. This results in zero model updates for all clients except the target victim. Therefore, the aggregated model updates are identical to the model update of the victim client, i.e. $\sum_{i=1}^{n}\delta_i=\mathbf{0}+\mathbf{0}+\cdots+\mathbf{0}+\delta_{target}=\delta_{target}$, exposing the model update of the target victim.

\begin{figure}[t]
    \centering
    \includegraphics[width=\linewidth]{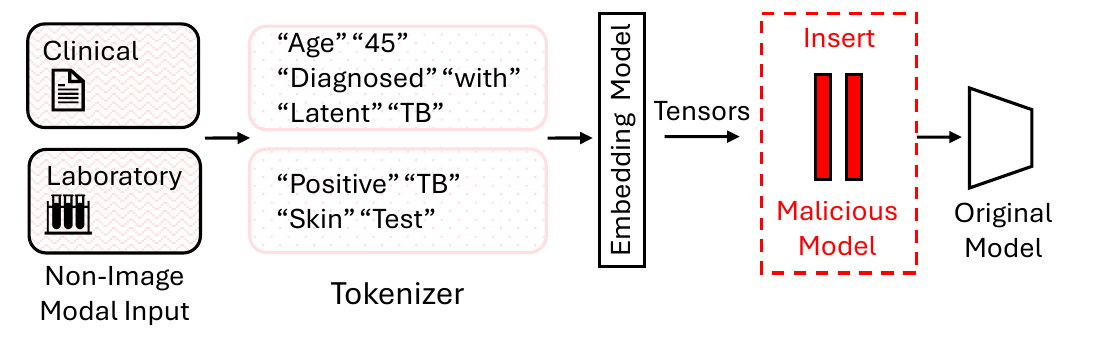}
    \caption{An example of how the medical text classification system works and where to insert the malicious modules.}
    \vspace{-10pt}
    \label{fig:text-flow}
\end{figure}

\paragraph{\textbf{Sample Reconstruction:}}
The sample reconstruction phase can be treated as the actual \textit{attack phase}, in which the parameter server disseminates the crafted global models $\hat{G}$ to all clients and recovers the local samples of the target victim $D_{target}$ according to the aggregated model updates $\sum_{i=1}^{n}\delta_i$ provided by all clients.

More specifically, with the help of the ``zero gradient'' module, the aggregated model update $\sum_{i=1}^{n}\delta_i$ the attacker obtains is identical to the model update of the victim client $\delta_{target}$, even though the SA protocols are in place. 
We further argue that the attacker can accurately estimate the gradients from the model update $\delta_{target}$ as it equals local iterations of the gradients \cite{mcmahan2017communication,shi2023scale}.
We define the gradients of the first two layers of the target victim as $g_{w_1}, g_{b_1}$ and $g_{w_2}, g_{b_2}$ respectively.
The attacker can calculate the following equation to reconstruct the input samples, for $l \in \{1,2,\cdots,k\}$:
\begin{equation}
    \label{reconsturction equation}
    \begin{aligned}
        (g_{w_1}^{(l+1)}-g_{w_1}^{(l)})/(g_{b_1}^{(l+1)}-g_{b_1}^{(l)}),
    \end{aligned}
\end{equation}
where specially we have $g_{w_1}^{(k+1)}$ and $g_{b_1}^{(k+1)}$ equal zero. 

\textbf{Analysis:} 
Equ. \ref{reconsturction equation} creates $k$ recovery bins to recover input images. Fortunately, when $k\ge m$, where $m$ refers to the size of the target dataset, each local training sample in the dataset will be \textit{perfectly recovered} within a certain bin ranging from $1$ to $k$. Here perfect recovery means that the inputs are analytically calculated through closed-form mathematical equations. A rigorous mathematical proof for this property is provided in the next section.
However, when $k<m$, there will be recovery conflicts, and some recovered samples are mixed with each other in certain bins, resulting in degraded recovery rates and quality. We argue this does not mean the total failure of the image reconstruction task. We will later demonstrate that in this scenario the attack performance gradually decreases and the attack remains to achieve decent performance when attack parameter $k$ is about the same scale as $m$.

We regard the attack parameter $k$ as the key factor that affects reconstruction performance.
Fortunately, from the attacker's perspective, this parameter is controlled and adjustable. The attacker can have a larger $k$ (i.e. craft larger linear layers) for large datasets and a smaller one for small datasets according to different attack scenarios to ensure that there are enough recovery bins for all samples. Regarding attack complexity, both the attack preparation and sample reconstruction phases only involve closed-form mathematical calculations that are super efficient to be conducted.

\subsection{Proof of Correctness}
Without the loss of generality, we use $x_p$ to denote the local input sample.
Considering the input $x_p$ falls in the $p^{th}$ largest bin, i.e. the brightness of $x_p$ denoted by $h(x_p)$ satisfies $h_p<h(x_p)<h_{p+1}$. We have the following equation holds:

\begin{equation}
\label{recovery equation}
\centering
    \begin{aligned}
        \frac{g_{w_1}^{(p+1)}-g_{w_1}^{(p)}}{g_{b_1}^{(p+1)}-g_{b_1}^{(p)}}
        &=\frac{\nabla_{w_1(p+1)}L-\nabla_{w_1(p)}L}{\nabla_{b_1(p+1)}L-\nabla_{b_1(p)}L}\\
        &=\frac{\frac{\partial L}{\partial y_{p+1}}\frac{\partial y_{(p+1)}}{\partial w_{1(p+1)}}-\frac{\partial L}{\partial y_{p}}\frac{\partial y_{(p)}}{\partial w_{1(p)}}}{\frac{\partial L}{\partial y_{p+1}}\frac{\partial y_{(p+1)}}{\partial b_{1(p+1)}}-\frac{\partial L}{\partial y_{p}}\frac{\partial y_{(p)}}{\partial b_{1(p)}}}\\
        &=\frac{\sum\limits_{v=1}^{p}\frac{\partial L}{\partial y_{p+1}}x_v-\sum\limits_{v=1}^{p-1}\frac{\partial L}{\partial y_{p}}x_v}{\sum\limits_{v=1}^{p}\frac{\partial L}{\partial y_{p+1}}-\sum\limits_{v=1}^{p-1}\frac{\partial L}{\partial y_{p}}}\\
        &=\frac{\frac{\partial L}{\partial y_{p}}x_p}{\frac{\partial L}{\partial y_{p}}}=x_p
    \end{aligned}
\end{equation}
where $L$ is the loss function, $y$ is the output of the first linear layer, and $\frac{\partial L}{\partial y_{p+1}}=\frac{\partial L}{\partial y_{p}}$ because we let the row vectors of the $w_2$ matrix identical. 

This equation implies $x_p$ is perfectly recovered from the gradients of the first linear layer. Because $\mathbf{H}$ covers the whole distribution range of the brightness feature, each input image will fall into one bin and thus can be recovered in this way as long as the image number is smaller than $k$.

\begin{figure*}[t]
    \centering
    \includegraphics[width=0.94\linewidth]{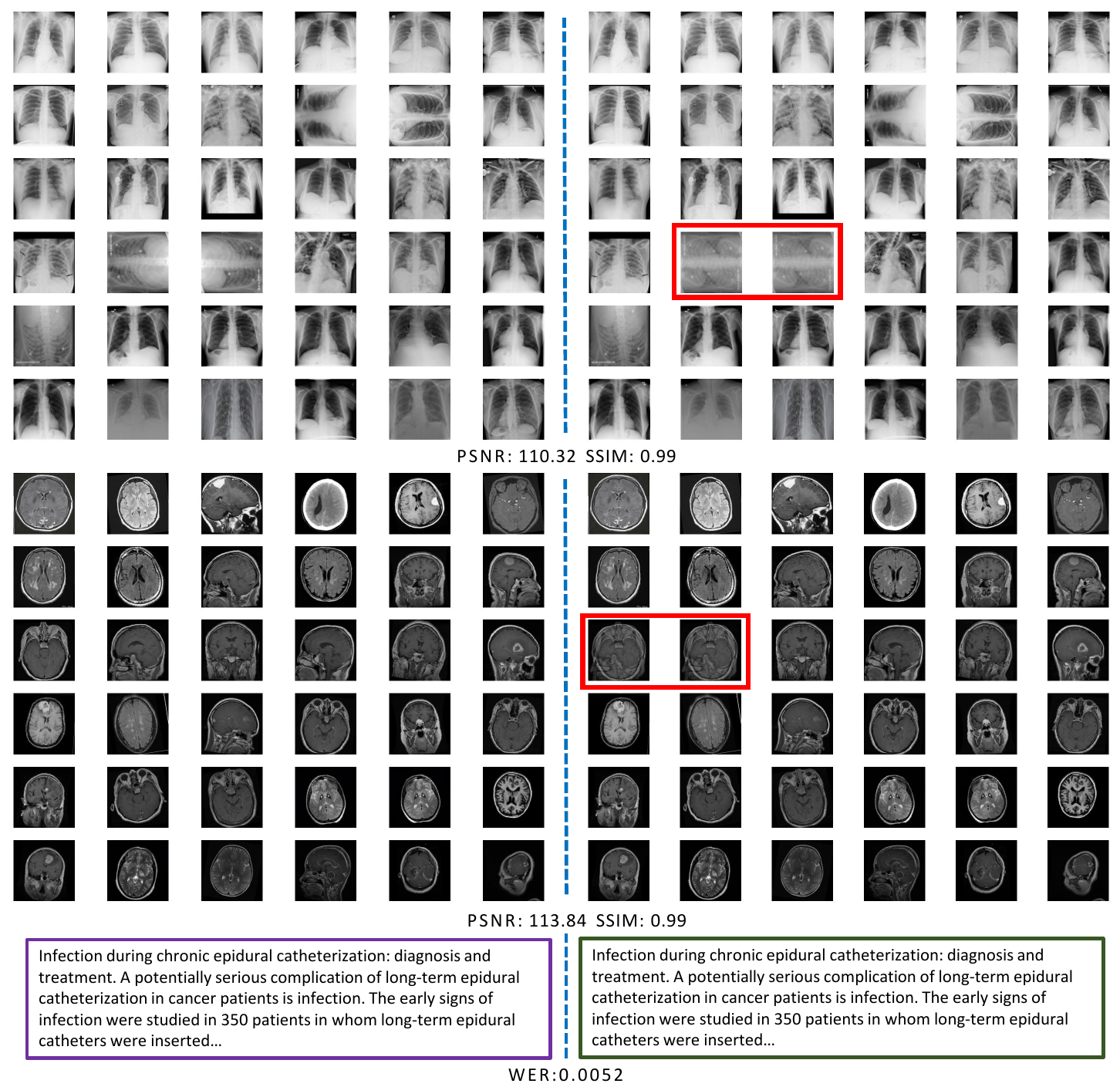}
    \vspace{-5pt}
    \caption{Recovered examples from the COVIDx CXR-4 dataset, Kaggle Brain Tumor MRI dataset, and MedAbstract dataset. The original images are on the left and the recovered ones are on the right. The text samples are truncated due to space limitations. Recovery failure samples are marked in red rectangles.
    }
    \vspace{-8pt}
    \label{fig:reco_covid_10}
\end{figure*}

\subsection{Text Data Reconstruction}

The medical text records are usually natural language words with discrete values within the input space. They have different mathematical properties from the medical images which are continuous pixel values in the input space. As a result, the medical natural language processing (NLP) models usually have different model architectures and workflows compared to their vision counterparts.
In Fig. \ref{fig:text-flow}, we demonstrate the architecture of a text classification model. More specifically, for an input sentence $x$, the model first converts it to a sequence of tokens $(t_1, t_2, \cdots, t_l)$, where $l$ refers to the maximum length of the sentence. Then the tokens are fed into the embedding model to be converted into the word embedding vectors, represented by $z=[e_1,e_2,\cdots,e_l]^T$, where all word embedding vectors $e_i$ have the same pre-defined embedding dimension. After that, the word embedding vectors are fed into the classification model to produce the output $o$.

Within this model architecture, our original attack strategy of placing the attack module in the front is not effective because the words are all discrete values and cannot be recovered in the same way as the continuous pixel values. 
Instead, we insert our attack module $L_{adv}$ including both the "linear leakage" module $L_{linear}$ for the target victim and the "zero gradient" module $L_{zero}$ for the others between the embedding layer and the classification model. In this way, according to our previous analysis, the attacker can successfully recover the embedding vectors $z=[e_1,e_2,\cdots,e_l]^T$ with the help of these attack modules. The next question is can the attacker further reverse the embedding vectors i.e. $e_i$ back to the tokens $t_i$? Fortunately, the answer is yes. This is because the word embedding layer maps the discrete tokens into continuous embedding vectors in the embedding space similar to a ``lookup table''; therefore, the attacker can simply select the word within the whole word vocabulary that minimizes the distance between the actual embedding vector and the calculated one to be the original token. In practice, we use a single Softmax layer to achieve this reverse function.


\begin{table*}[t]
\caption{The reconstruction performance of \sysname over different datasets and reconstruction batch sizes. The rate (sample recovery rate) is on a scale of 1.00}
\centering
\begin{tabular}{ccccccc} 
\hline
\textbf{Batch Size}\qquad &  \textbf{Dataset}\qquad  & \textbf{Pixel Size}\qquad & \textbf{Rate}\qquad & \textbf{PSNR}\qquad & \textbf{SSIM}\qquad & \textbf{Time (in sec)}\qquad \\
\hline
\multirow{3}{*}{100} & ChestMNIST(pneumonia) & 28x28 & 1.0  & 112.574& 0.99 & 0.742\\ 
    & COVIDx CXR-4 & 224x224 & 0.951 & 120.795 & 0.99 & 6.022\\
    & Kaggle Brain Tumor MRI & 224x224 & 0.962 & 107.783 & 0.99 & 6.308 \\
\hline
\multirow{3}{*}{200} & ChestMNIST(pneumonia) & 28x28 & 0.960 & 102.722& 0.99& 0.936\\ 
    & COVIDx CXR-4 & 224x224 & 0.891 & 114.982 & 0.99 & 7.003\\
    & Kaggle Brain Tumor MRI & 224x224 & 0.870 & 98.6421 & 0.99 & 6.738
    \\
\hline
\multirow{3}{*}{300} & ChestMNIST(pneumonia) & 28x28 & 0.957& 97.405& 0.99& 0.95\\ 
    & COVIDx CXR-4 & 224x224 & 0.880 & 105.123 & 0.99 & 8.121\\
    & Kaggle Brain Tumor MRI & 224x224 & 0.845 & 96.722 & 0.99 & 7.181
    \\
\hline
\multirow{3}{*}{400} & ChestMNIST(pneumonia) & 28x28 & 0.955& 93.713& 0.99& 1.020\\ 
    & COVIDx CXR-4 & 224x224 & 0.864 & 97.301 & 0.99 & 8.762\\
    & Kaggle Brain Tumor MRI & 224x224 & 0.804 & 93.179 & 0.99 & 8.042
    \\
\hline
\multirow{3}{*}{500} & ChestMNIST(pneumonia) & 28x28 & 0.964& 87.019 & 0.99& 1.016\\ 
    & COVIDx CXR-4 & 224x224 & 0.810 & 95.864 & 0.99 & 9.763\\
    & Kaggle Brain Tumor MRI & 224x224 & 0.796 & 92.954 & 0.99 & 8.767
    \\
\hline
\end{tabular}
\vspace{-5pt}
\label{tab:compare_dataset_perform}
\end{table*}

\textbf{Analysis}: Similar to the image recovery task, the performance of the text recovery task is largely affected by the number of recovery bins $k$. We assume during the FL training process, $m$ text data samples are organized in a batch with each one having a maximum length of $l$ words. We clarify that the maximum sentence length $l$ does not affect the recovery performance as long as it is not super large (e.g. $10^4$), which we will later justify in the experiment section. The recovery performance is still largely determined by the relationship between $m$ and $k$. When $k\ge m$, each embedding vector is perfectly recovered within one bin, which further leads to excellent text recovery performance. But when $k<m$, there will be recovery collisions within certain bins, and the text recovery performance drops. 

\section{Evaluation} 

\begin{figure*}[t]
\centering
    \subfigure[PSNR Scores.]{\includegraphics[width=0.32\textwidth]{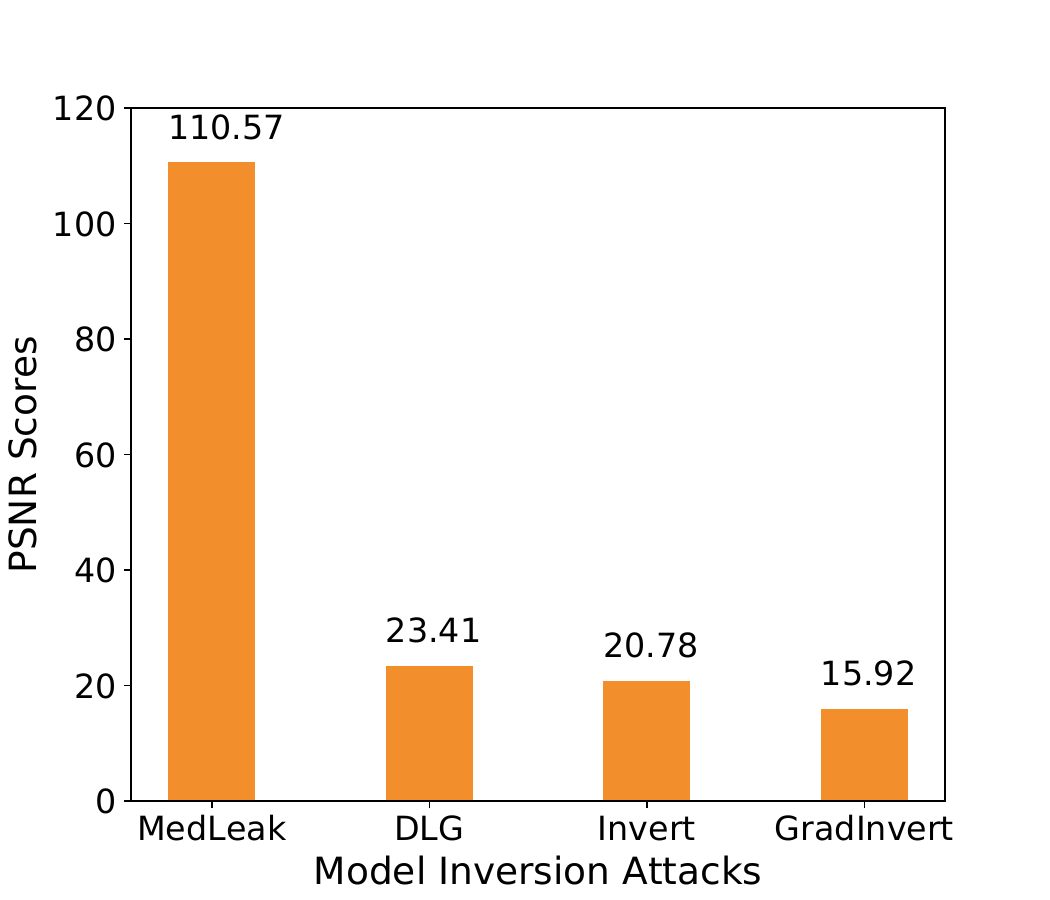}}
    \subfigure[SSIM Scores.]{\includegraphics[width=0.32\textwidth]{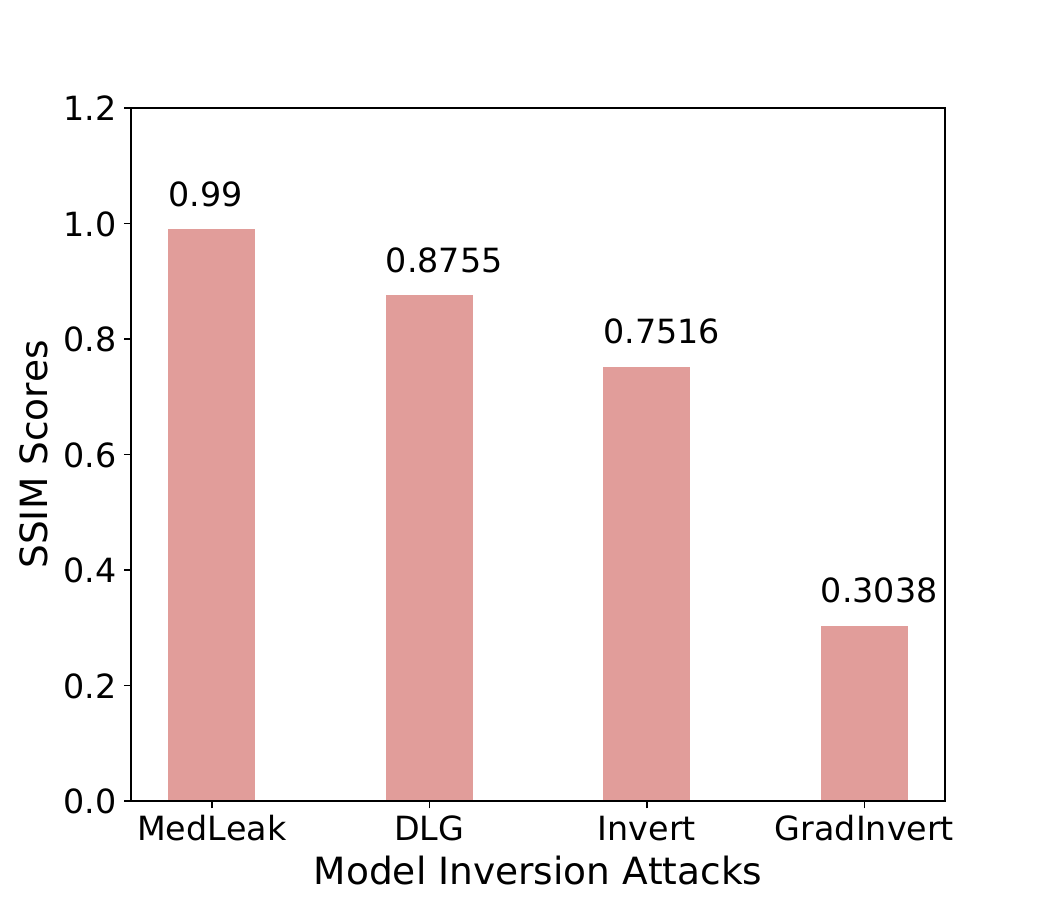}}
    \subfigure[Attack Time (Seconds).]{\includegraphics[width=0.32\textwidth]{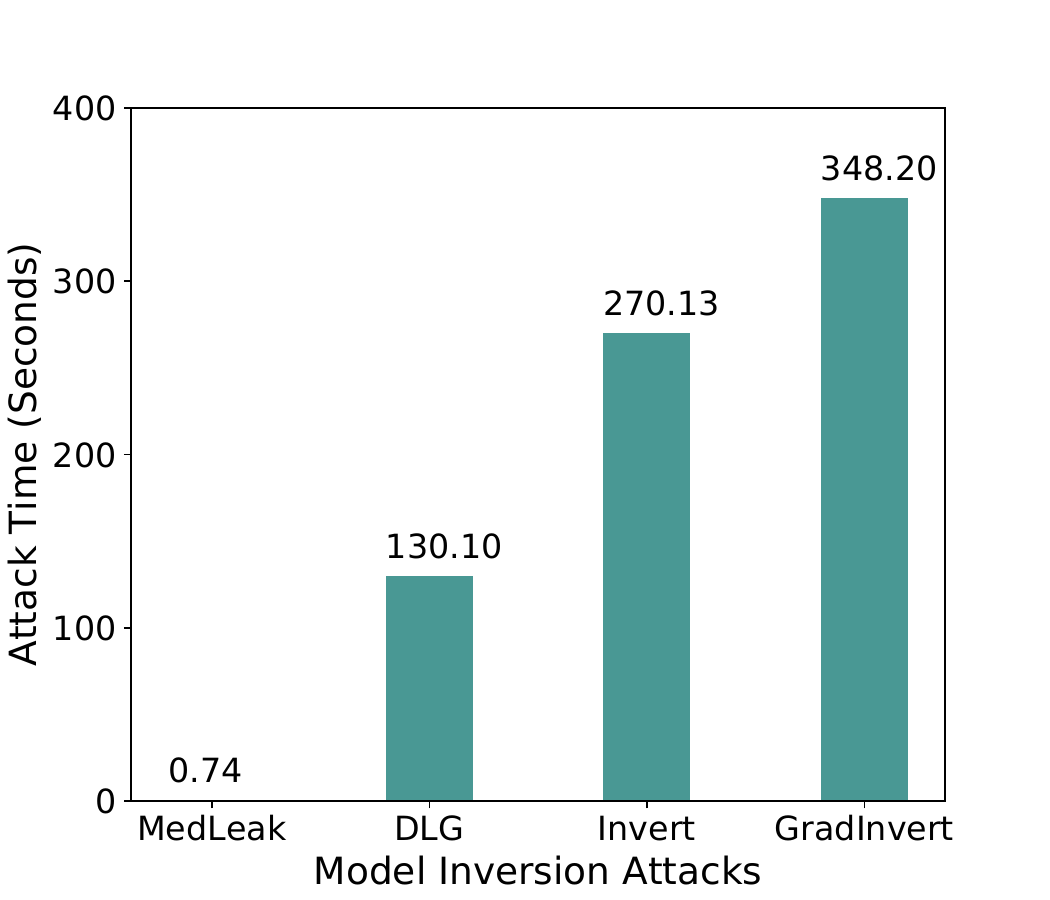}}
    \vspace{-8pt}
    \caption{The attack performance comparison between \sysname with other model inversion attacks.}
     \vspace{-8pt}
    \label{fig:Benchmark comparison} 
\end{figure*}

\subsection{Experimental Settings} 
We implemented \sysname on the PyTorch platform. We ran all the experiments on a server equipped with an Intel Core i7-12700K CPU 3.60GHzX12, one NVIDIA GeForce RTX 3080 Ti GPU, and Ubuntu 20.04.6 LTS. 

We considered the FL system to have 5 clients with one of them being the attack target per training round. We assumed each client would perform 5 local iterations before generating the individual model updates. We randomly selected $10\%$ of the training set as the auxiliary dataset and aimed to recover samples in the test set, which have \textit{no intersection} with the auxiliary dataset. We assumed the test set is partitioned and owned by the 5 clients locally to serve as the local datasets. 
For the defense mechanism, we considered the system to be protected by the SA protocol in \cite{bonawitz2017practical}. Therefore, we cannot get access to the individual model updates (which have been cryptographically masked) and we launched our attack solely based on the aggregated model updates. Note that our attack can not only break the SA protocol in \cite{bonawitz2017practical} because \sysname breaks the abstract aggregation function of the SA protocols regardless of their implementation details. All SA mechanisms realizing the function can be broken, including \cite{bell2020secure, choi2020communication, guo2020v, kadhe2020fastsecagg,pillutla2022robust, burkhalter2021rofl}.

For the image recovery task, we chose the ChestMNIST dataset from the MedMNIST package \cite{medmnistv2}, the COVIDx CXR-4 dataset \cite{Wang2020}, and the Kaggle Brian Tumor MRI dataset \cite{brian-tumor-mri} as our experiment datasets. The ChestMNIST dataset comprises frontal view X-ray images (1×28×28) of 30805 unique patients with 14 disease labels and we selected data samples related to pneumonia to conduct our experiments, including 78468 training samples and 22433 testing samples. The COVIDx CXR-4 dataset also consists of frontal-view chest X-ray images with higher dimensions (resized to 1×224×224) and labels about whether the patient is COVID-positive. The training set contains 67863 images and the testing set contains 8482 images. The Kaggle Brain Tumor MRI Dataset contains 7023 images of human brain MRI images which are classified into 4 classes: glioma - meningioma - no tumor and pituitary. Its training set contains 5712 images and the testing set contains 1311 images. 

We used four evaluation metrics including the recovery rate, the attack time, the peak signal-to-noise ratio (PSNR) score, and the structural similarity index measure (SSIM) score, following the convention of the existing works \cite{zhu2019deep,zhao2020idlg,geiping2020inverting,yin2021see,fowl2021robbing} to evaluate our attack over the image recovery task. More specifically, the successful recovery of samples was measured by observing the PSNR and SSIM scores between the original input samples and the reconstructed ones and checking whether those scores exceed a certain threshold $th$. In our work, we chose $th = 20$ for PSNR and $th = 0.9$ for SSIM because these thresholds are enough to ensure that the recovered images are visually clear for the attacker to extract all meaningful content from them. 

For the text recovery task, we chose the MedAbstract dataset \cite{MedAbstract} as our experiment dataset. The MedAbstract data set consists of 14438 medical abstracts describing the patients' health conditions with each one consisting of a few hundred words. The patients' conditions are classified into 5 different classes including digestive system diseases, cardiovascular diseases, neoplasms, nervous system diseases, and general pathological conditions. 

We used three evaluation metrics including the recovery rate, the word error rate (WER), and the attack time to evaluate \sysname's performance over the text recovery task. Similar to the vision task, the recovery rate was defined as the ratio between the number of successfully recovered text samples achieving WERs lower than threshold 0.05 and the total sample number. Meanwhile, the WER was calculated as the portion of recovery failure words in text samples. In our experiment, we focused on the WER of the successfully recovered samples to evaluate their recovery quality.

\begin{table*}[t]
\caption{Downstream binary classification task on the COVID dataset with a pre-trained (with CXR-3 dataset) ViT model. TPR: True Positive Rate, TNR: True Negative Rate, ACC: Accuracy, AUC: Area Under Receiver Operating Characteristic curve, AUPR: Area Under Precision Recall curve.}
\centering
\begin{tabular}{l|l|lllll}
\hline 
\textbf{Model}\qquad & \textbf{Image}\qquad & \textbf{AUPR}\qquad & \textbf{TNR}\qquad & \textbf{TPR}\qquad & \textbf{ACC}\qquad & \textbf{AUROC}\qquad \\
\hline 
\multirow{2}{*}{ViT-S (SSL)}& Original & 0.937 & 0.800  & 0.857 & 0.829 & 0.905 \\
& Recovered & 0.921 & 0.900  & 0.710   & 0.805 & 0.919  \\
\hline 
\multirow{2}{*}{ViT-S (Fine-tuned)} & Original & 0.974 & 0.970 & 0.930 & 0.950 & 0.969 \\
& Recovered & 0.965 & 0.886 & 0.938 & 0.912 & 0.966 \\
\hline
\end{tabular}
\vspace{-5pt}
\label{tab:vit_perform}
\end{table*}

\subsection{Image Reconstruction Results}

In Tab. \ref{tab:compare_dataset_perform}, we demonstrate the performance of \sysname over different recovery batch sizes (i.e. the number of samples recovered simultaneously held by the target victim) ranging from 100 to 500 images. 
We can observe that both the recovery rate (i.e. the ratio of successfully recovered images) and the quantitative scores (i.e. the PSNR and SSIM scores) decrease when the recovered batch size increases. This is expected because the larger the recovery batch size, the more difficult the recovery task to conduct. 
But in general, \sysname achieves high recovery rates ($>0.8$) and quantitative scores (PSNR$>80$ and SSIM$>0.9$) for all three datasets under all recovery batch sizes. Particularly, the SSIM scores (ranging from 0 to 1) remain to be 0.99 for all settings, because of the recovery excellency.  This can be further verified by the recovered samples we visualized in Fig. \ref{fig:reco_covid_10}, in which we plot the original images on the left and the recovered ones on the right. We find that the recovered images are of high quality and cannot be visually distinguished from the original ones, even for some detailed small marks and notations. In Tab. \ref{tab:compare_dataset_perform} we also demonstrate the attack time (in seconds). We find that the attack time is monotonically increasing with respect to the recovery batch size. For the largest batch size (i.e. 500 images) over the complex COVIDx CXR-4 dataset, it only takes the attacker less than 10 seconds to fulfill the recovery task, indicating that \sysname is very effective.

\subsection{Benchmark Comparison}

We compared \sysname's attack performance with three optimiza\/tion-based model inversion attacks (MIAs) including the DLG/iDLG \cite{zhu2019deep,zhao2020idlg}, InvertGradient \cite{geiping2020inverting}, and GradInversion \cite{yin2021see} attacks (denoted as DLG, Invert, and GradInvert respectively) over the MedMNIST dataset for one small batch of input images. 
We only focused on the image recovery task because all the existing attacks cannot be adapted to the text recovery task. 
We compared the PSNR scores, SSIM scores, and the attack time between the three attacks and our work. The results are demonstrated in Fig. \ref{fig:Benchmark comparison}. We can find that our attack achieves much better PSNR scores and SSIM scores than the existing MIAs, indicating that our attack can reconstruct samples with better quality. At the same time, our attack consumes significantly less time than the current MIAs. Particularly, the existing attacks consume a few hundred seconds to reconstruct one batch of samples, while our attack only requires less than one second, which reduces the current cost by two orders of magnitude. The reason why our attack is much more efficient is that our attack only involves closed-form mathematical calculations while the other three attacks require costly iterative-based optimization methods. However, there is no free lunch and we clarify that the three benchmark works adopt an honest-but-curious attack model, which does not allow the attacker to modify the model parameters and architecture as we did.

\begin{table*}[t]
\caption{The text reconstruction performance of \sysname over a different number of text samples and experiment settings. The rate (sample recovery rate) is on a scale of 1.00. The ``Embed Dim'' refers to the embedding dimension.}
\centering
\begin{tabular}{cccccc} 
\hline
\textbf{Text Num}\qquad &  \textbf{Max Length}\qquad & \textbf{Embed Dim}\qquad & \textbf{Rate}\qquad & \textbf{WER}\qquad & \textbf{Time (in sec)}\qquad \\
\hline
\multirow{2}{*}{20} & 200 words & 64  & 0.9375 & 0.0004 & 0.2149 \\
& 300 words & 64  &  0.9669 & 0.0009 & 0.3057 \\  

\hline
\multirow{2}{*}{40}  & 200 words  & 64& 0.9212 & 0.0004 & 0.416 \\ 
& 300 words & 64  & 0.9153 & 0.0018 & 0.6023 \\  
\hline
\multirow{2}{*}{60}  & 200 words & 64& 0.8729 & 0.0005 & 0.6341 \\
& 300 words & 64  & 0.9083 & 0.002 & 0.9192\\  

\hline
\multirow{2}{*}{80} & 200 words & 64& 0.8228 & 0.0023 & 0.8331 \\ 
& 300 words & 64  & 0.8540 & 0.0051 & 1.2308\\  

\hline
\multirow{2}{*}{100}  & 200 words & 64& 0.755 & 0.0047 & 1.058\\ 
& 300 words & 64  & 0.7585 & 0.0052 & 1.514 \\  

\hline
\end{tabular}
\label{tab:text_performance}
\end{table*}

\begin{figure*}[t]
\centering
    \subfigure[Recovery bins factor.]{\includegraphics[width=0.238\textwidth]{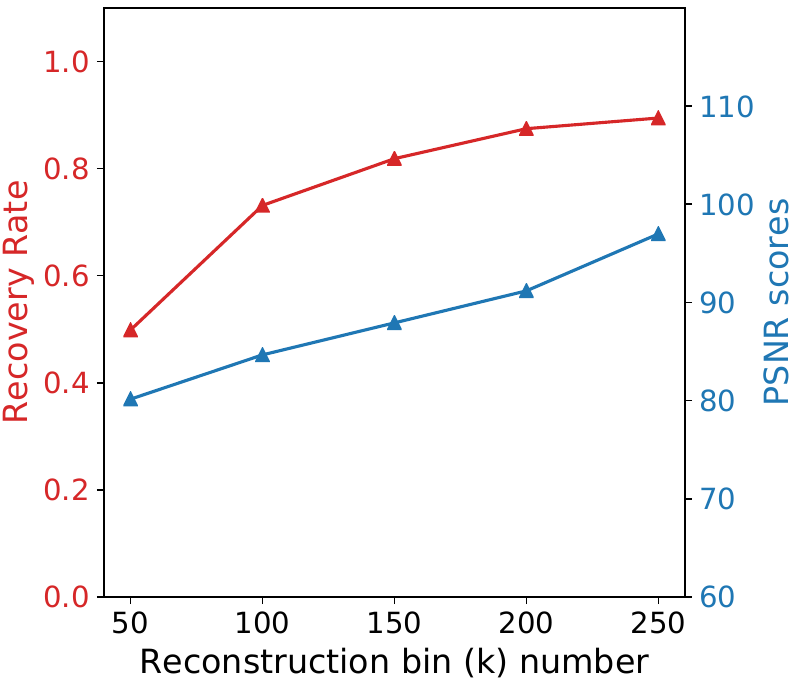}}
    \subfigure[Client number factor.]{\includegraphics[width=0.238\textwidth]{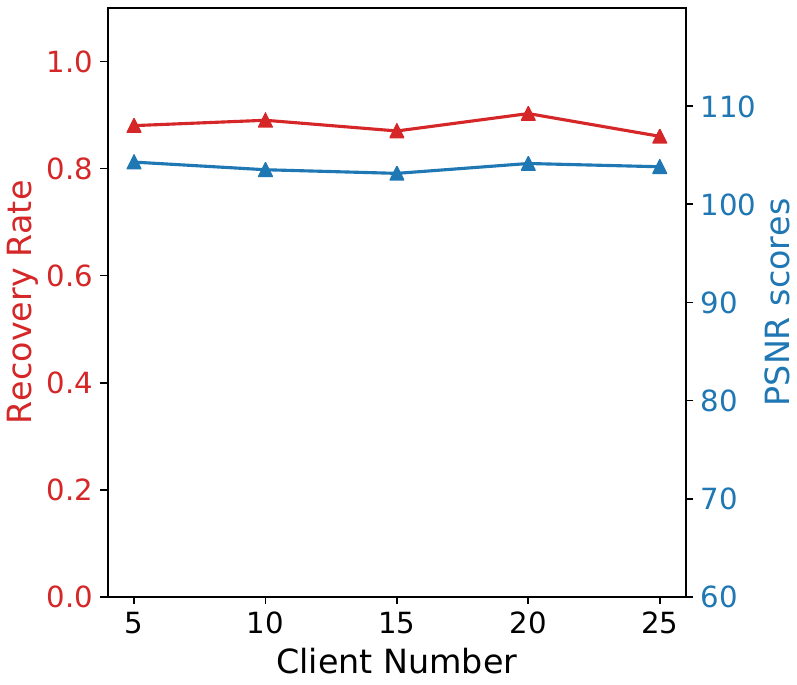}}
    \subfigure[Local training epochs factor.]{\includegraphics[width=0.238\textwidth]{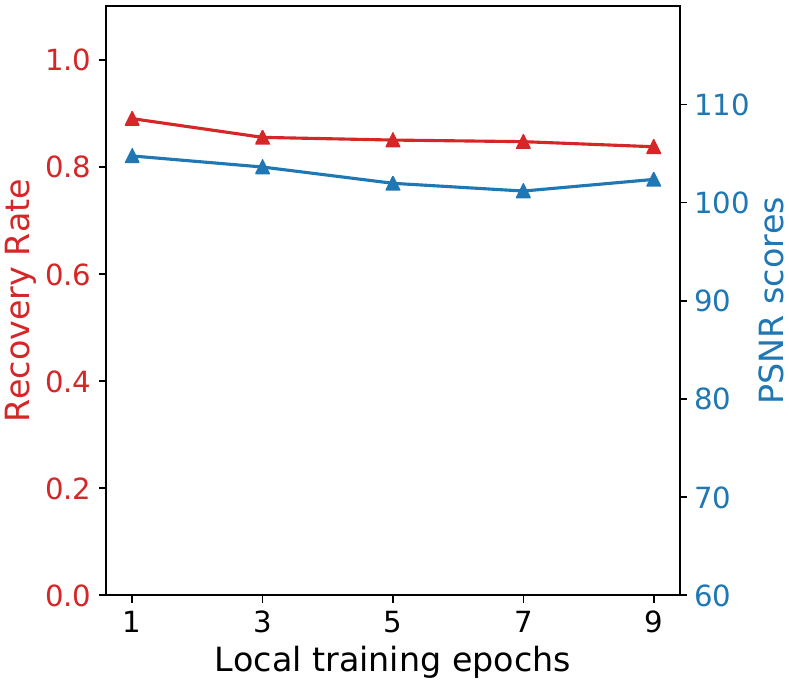}}
    \subfigure[Non-iid factor.]{\includegraphics[width=0.238\textwidth]{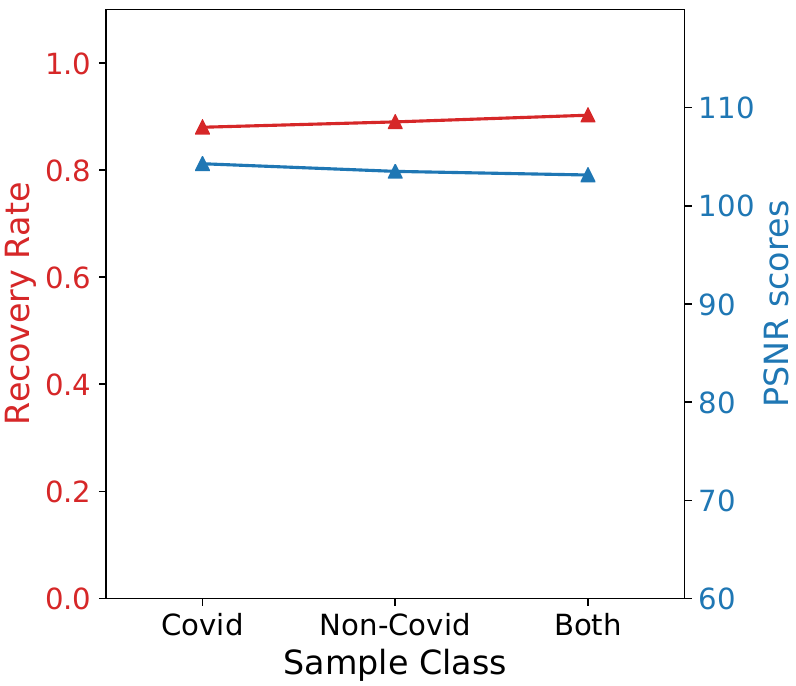}}
    \vspace{-8pt}
    \caption{The recover performance of \sysname over different practical attack factors.}
    \vspace{-8pt}
    \label{fig:Performance Factor} 
\end{figure*}

\subsection{Vision Downstream Tasks}
To further evaluate the performance of our attack on clinically relevant downstream tasks, we performed a binary disease classification (the detection of COVID-19) task on both the recovered samples and the actual samples. We used the state-of-the-art vision transformer model (ViT-S) (embedding size=368, number of heads=6, 22M parameters) pre-trained by self-supervised learning (SSL) technique on 30k COVIDx CXR-3 samples and fine-tuned on the RSNA-RICORD part of the dataset to perform the classification task \cite{anwar2023spcxr} and evaluated it on the COVIDx CXR-4 dataset. We demonstrate the performance in Tab. \ref{tab:vit_perform}. We use widely used machine learning metrics to evaluate the classification performance and we find that the recovered images achieve nearly the same performance as the original ones.
This shows that our reconstruction process is highly successful in keeping all semantic meaning within the images and the reconstructed images can be used to perform any potential clinical analysis, which further indicates the severity of the privacy threat imposed by our attack. We consider it a very practical attack scenario for a curious party, which either can be the medical federated learning's participants or a third-party service provider (who provides the necessary platform, computation resource, and other FL infrastructures) to launch our attack to first reconstruct the sensitive medical images and then feed them to certain downstream analysis tasks to obtain further information about of the victims.

\subsection{Text Reconstruction Results}
In Tab. \ref{tab:text_performance}, we demonstrate \sysname's performance on recovering medical text data under different batch sizes (ranging from 20 to 100) and data settings.
More specifically, ``text num" refers to the number of text samples recovered simultaneously in one batch, and ``max length'' refers to the maximum length of the text contents in the number of words. In our experiment, we fixed the length of each text sample for processing convenience. We truncated the samples when their lengths were longer than the maximum length and padded them when they were shorter. We also fixed the embedding dimension as the commonly used value 64. 
From the result, we can find that in general, \sysname achieves decent text recovery performance under different settings to obtain high recovery rates ($>0.75$), low WER scores ($<0.006$), and short execution time within a few seconds. We observe that when the recovery batch size increases, the recovery rate decreases accordingly, meaning that recovering a larger batch of samples is more difficult. 
By comparing \sysname's performance under different sample lengths (i.e. 200 v.s. 300 words), we further observe that a larger sample length triggers longer attack time, which approximately follows a linear relationship with the sample length. However, a larger sample length does not trigger any recovery performance drop. We still observe decent and stable recovery rates and WER scores when the sample length increases. In fact, the recovery rate even increases when the sample length is larger. This may be because longer samples maintain more semantic information and can be better separated and recovered. 

\subsection{Performance Affecting Factors}

In this section, we investigate \sysname's performance under different FL settings. We implemented the attack on the COVID CXR-4 dataset and evaluated the recovery rates and PSNR scores considering the following 4 factors: recovery bin number, client number, local training epoch, and non-iid data settings. 

\textbf{Recovery Bin Number:} 
As we have discussed in Section \ref{Method}. The recovery bin size $k$ (i.e. neuron number of the first linear layer) can significantly affect the performance of \sysname. 
In the experiment, we fixed the reconstruction batch size to 100 and changed the recovery bin $k$ size from 50 to 250. We demonstrate the results in Fig. \ref{fig:Performance Factor}(a). We find that both the recovery rates and PSNR scores are monotonically increasing with $k$ and obtain relatively high quantitative scores when $k\ge m$. This is consistent with our theoretical analysis as more bins (larger $k$) will decrease the recovery conflict probability and increase the recovery success rates. In practice, the attacker can adjust the bin size according to local sample size to obtain decent attack performance.

\textbf{Client Number:} We increased the client number from 5 to 25 in our experiment. The results are shown in Fig. \ref{fig:Performance Factor}(b). We find that the attack performance is not affected by the FL client number. This indicates that our dis-aggregation attack phase is highly successful and the attacker can always obtain accurate model updates of the single victim client.

\textbf{Local Training Epoch:} We increased the local training epochs of the FL clients from 1 to 9. The results are demonstrated in Fig. \ref{fig:Performance Factor}(c). From the results we only observe very slight performance degradation when the FL clients conduct more local training epochs, demonstrating that our attack can be applied to both FedSGD and FedAVG settings.

\textbf{Non-iid Data Distribution:} We changed the class of samples held by the victim clients in our experiment including having COVID-only, non-COVID-only, and both type of samples. The results are shown in Fig. \ref{fig:Performance Factor}(d). We observe that the attack performance is not affected by the type of samples held by the victim client, indicating that \sysname are applicable to both iid and non-iid settings.

\section{Discussion and Future Work}

\textbf{Auxiliary Dataset:} Using an auxiliary dataset is a commonly used prerequisite for state-of-the-art MIAs \cite{pasquini2022eluding,fowl2021robbing,zhao2023loki, shi2023scale}. 
Our attack takes the same assumption and its performance is also affected by the number and quality of samples in the auxiliary data $D_{aux}$. 
In the ideal case, the auxiliary dataset shall have the same data distribution as the target victim's local dataset, or the auxiliary dataset is more representative. It may be a challenge for the general vision tasks to find such a representative auxiliary dataset. However, for the medical data, this does not pose a critical barrier, since the radiology data (such as the CT scans) of humans are acquired in similar data format with common anatomical features. The attacker can obtain representative datasets released for research purposes in public domains. Moreover, because the attacker is a participant in the FL system, we consider they may even collude with others to obtain this auxiliary dataset.


\textbf{Defense Mechanisms:} 
An intuitive yet effective defense against our attack is for clients to proactively verify the consistency of model parameters and architectures during the FL training process, rather than blindly trust the privacy guarantees of FL systems and place full confidence in FL service providers—a trust that is prevalent in the current medical FL systems. However, we argue that our attack can be initiated during the initial training rounds, or even in the very first round, to maintain its stealthiness. This is because, in the early stages of training, when everything is randomly initialized, it becomes challenging for defenders to distinguish between malicious activities and benign patterns that arise from random initialization.

Another potential defense mechanism comes from the data synthesis perspective, leveraging the inherent bottleneck of \sysname.
Our comprehensive analysis reveals that the recovery bin size, denoted as $k$, plays a crucial role in affecting the attack performance. Specifically, when the sample size $m$ is much larger than $k$, \sysname's performance drops significantly.
Based on this finding, the clients can generate large \textit{mask sets} containing many images but not exposing anything related to the original private local samples to crowd the recovery bins. When the size of the mask set is large enough, there will be a lot of collisions within the bins, and the attacker can hardly recover anything. This defense strategy can also be employed for other attacks that face similar performance bottlenecks such as \cite{fowl2021robbing,zhao2023loki}. However, the primary design challenge for such defense is to avoid causing any performance degradation when these mask sets are involved in the FL training process.


\textbf{More than Linear Leakage:} 
In this work, we leverage the fundamental ``linear leakage'' primitive as a powerful mathematical tool to help us accomplish our attack. However, we acknowledge that other types of model components such as the vision transformer \cite{lu2022april}, and convolutional layers \cite{zhao2023loki} may also be exploited to reverse the model updates and leak private training data. These model components are integral to popular machine-learning models and can be exploited to launch the attack without the need to insert any additional malicious modules. This would make the attack more stealthy and adaptable for different victim models.
However, designing such an analytical gradient reverse method is non-trivial, and we intend to explore such designs in future work.


\section{Conclusion}

In this paper, we present \sysname--a novel MIA that targets current FL systems designed for healthcare applications using sensitive patient data. 
\sysname can accurately and efficiently recover local training samples at a clinical site, 
resulting in unwanted leakage of private patient information. To achieve this, \sysname requires the parameter server (i.e., the attacker) to actively craft additional adversarial attack modules before the global models. These adversarial models are designed with the mathematical guarantee to effectively break the secure aggregation protocol and efficiently recover hundreds of samples in a batch without relying on costly optimization methods when they are sent to the clients. We customize \sysname to recover both image and textual data records, as clinical data usually comprises both types of samples, extending its applicability to wider healthcare-related FL systems.
We implement \sysname on multiple medical images and text datasets and our results highlight \sysname's excellent attack performance under various real-world settings.
Our attack exposes a practical vulnerability of the current medical FL systems, prompting the community to reconsider the privacy guarantees of these systems and to develop effective defenses against such advanced MIAs.

\section*{Acknowledgments}

This work was supported in part by the Office of Naval Research under grants N00014-24-1-2730 and N00014-19-1-2621, the National Science Foundation under grants 2312447, 2247560, 2154929, 2332675, and 2235232, and Children's National Hospital, the Sanghani Center for AI and Data Analytics, and the Fralin Biomedical Research Institute at Virginia Tech. 
\bibliographystyle{ACM-Reference-Format}
\bibliography{reference}
\end{document}